\documentclass{article}

\usepackage{arxiv}

\usepackage[utf8]{inputenc} 
\usepackage[T1]{fontenc}    
\usepackage{hyperref}       
\usepackage{url}            
\usepackage{booktabs}       
\usepackage{amsfonts}       
\usepackage{nicefrac}       
\usepackage{microtype}      
\usepackage{lipsum}		
\usepackage{graphicx}
\usepackage{natbib}
\usepackage{doi}

\usepackage{multirow}
\usepackage{amssymb}
\usepackage{amsmath}
\usepackage{graphicx} 
\usepackage[table,xcdraw]{xcolor}

\title{CIEC: Coupling Implicit and Explicit Cues for Multimodal Weakly Supervised Manipulation Localization}

\author{ 
	Xinquan Yu \\
	School of Computer Science and Engineering, MoE Key Laboratory of Information Technology, \\ Guangdong Province Key Laboratory of Information Security Technology, Sun Yat-sen University, \\
	Guangzhou 510006, China \\
	\texttt{yuxq28@mail2.sysu.edu.cn} \\
	\AND
	Wei Lu \\
	School of Computer Science and Engineering, MoE Key Laboratory of Information Technology, \\ Guangdong Province Key Laboratory of Information Security Technology, Sun Yat-sen University, \\
	Guangzhou 510006, China \\
	\texttt{luwei3@mail.sysu.edu.cn} \\
	\AND
	Xiangyang Luo \\
	State Key Laboratory of Mathematical Engineering and Advanced Computing, \\
	Zhengzhou 450002, China \\
	\texttt{luoxy\_ieu@sina.com} \\
	\AND
	Rui Yang \\
	Alibaba Group, \\
	Hangzhou, China \\
	\texttt{duming.yr@alibaba-inc.com} \\
}

\renewcommand{\shorttitle}{\textit{arXiv} Template}

\begin{document}
\maketitle

\begin{abstract}
	To mitigate the threat of misinformation, multimodal manipulation localization has garnered growing attention.
	Consider that current methods rely on costly and time-consuming fine-grained annotations, such as patch/token-level annotations.
	This paper proposes a novel framework named Coupling Implicit and Explicit Cues (CIEC), which aims to achieve multimodal weakly-supervised manipulation localization for image-text pairs utilizing only coarse-grained image/sentence-level annotations.
	It comprises two branches, image-based and text-based weakly-supervised localization.
	For the former, we devise the Textual-guidance Refine Patch Selection (TRPS) module. It integrates forgery cues from both visual and textual perspectives to lock onto suspicious regions aided by spatial priors. Followed by the background silencing and spatial contrast constraints to suppress interference from irrelevant areas.
	For the latter, we devise the Visual-deviation Calibrated Token Grounding (VCTG) module. It focuses on meaningful content words and leverages relative visual bias to assist token localization. Followed by the asymmetric sparse and semantic consistency constraints to mitigate label noise and ensure reliability.
	Extensive experiments demonstrate the effectiveness of our CIEC, yielding results comparable to fully supervised methods on several evaluation metrics.
\end{abstract}

\section{Introduction}
Nowadays, the rapid advancement of AIGC has significantly reduced the cost of creating realistic content \cite{radford2019language, patashnik2021styleclip, kang2023scaling, liu2024deepseek}, enriching people's mental lives. However, some individuals intentionally create multimodal misinformation to attract attention, which severely affects people's perception of content authenticity \cite{lazer2018science, wu2019misinformation, yu2025racmc}. 

\begin{figure}[t]
	\centering
	{\includegraphics[width=0.7\linewidth]{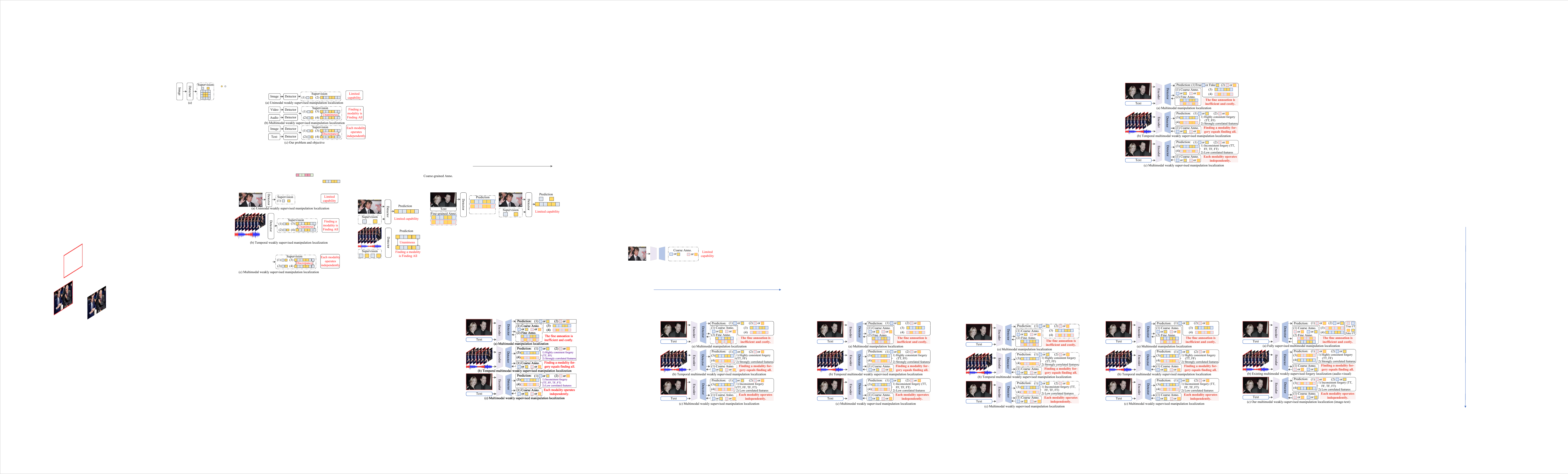}}
	\caption{Comparison of different multimodal manipulation localization paradigms. (a) Fully supervised localization relies on expensive fine-grained annotations. (b) Existing multimodal weakly-supervised forgery localization (audio-visual) only considers scenarios where both modalities are true or fake, benefiting from high cross-modal consistency. (c) Our task (image-text) encounters the dual challenge of cross-modal inconsistency forgery (e.g., TF, FT) and stronger modality independence.
	}
	\label{fig1}
\end{figure}

To alleviate this, scholars have researched the task of detecting and localizing multimodal misinformation, e.g., HAMMER \cite{shao2023detecting} constructs a first large-scale multimodal manipulation localization dataset named DGM$^4$ for image-text modality, and achieves multimodal localization through a two-stage shallow-to-deep progressive reasoning approach. On this foundation, HAMMER++ \cite{shao2024detecting} refines cross-modal alignment to enhance performance.
Given that forged content may impact cross-modal semantics, ViKI\cite{li2024towards} performs feature alignment only on fully authentic multimodal content.
Unlike the previous two-stage reasoning framework, UFAFormer \cite{liu2024unified} proposes a unified detection framework assisted by the frequency domain.
By decoupling image features, IDseq \cite{liu2025idseq} permits only authentic image features to assist text localization, thereby reducing erroneous guidance from fake image regions.
ASAP \cite{zhang2025asap} utilizes the Large Language Model (LLM) for the first time to generate auxiliary textual information for image and text modalities, thereby facilitating cross-modal alignment.
CSCL \cite{li2025unleashing} employs cascaded context-consistency and semantic-consistency decoders to enhance its fine-grained perception of forgery.

Although existing methods have achieved promising results, they rely on expensive pixel-level and token-level annotations (in Figure \ref{fig1}(a)), which are inefficient and costly to acquire at scale.
To alleviate this burden, researchers have explored Weakly-supervised Forgery Localization (WFL) methods, but these primarily target unimodal data such as image \cite{pathak2014fully, araslanov2020single} and audio \cite{wu2025weakly}. Recently, MDP \cite{xu2025multimodal} investigate weakly-supervised forgery localization for temporal multimodal data. However, in this scenario, multimodal forgeries exhibit strong temporal correlation and forgery consistency (in Figure \ref{fig1}(b)). In other words, when the forgery of one modality is accurately located, the forgery of another modality will naturally be located.
However, this logic does not apply to complex and static image-text scenarios (in Figure \ref{fig1}(c)), where the forgery locations of image and text are highly independent.

To this end, we propose an image-text multimodal weakly-supervised manipulation localization framework, named CIEC, to couple implicit and explicit cues. 
Specifically, we first utilize the Cross-modal Feature Alignment (CFA) module to bridge the heterogeneity between two modalities. It adopts a dual-stream multi-layer interaction architecture to separately process image and text modalities. 
Second, we devise the Textual-guidance Refine Patch Selection (TRPS) module to achieve image WFL, which primarily leverages pre-selected candidate boxes as spatial priors to alleviate the limitations of global searches. In more detail, it extracts forgery clues from both implicit visual and explicit textual perspectives to lock the most suspicious regions. Then, it utilizes background silence constraints to suppress irrelevant responses from non-targeted regions, while further enhancing the discernibility between true and fake features through spatial contrast forgery enhancement constraints.
Finally, we design the Visual-deviation Calibrated Token Grounding (VCTG) module for text WFL, which primarily focuses on meaningful content words. In more detail, it employs the 1D convolution to capture intrinsic textual forgery traces and mines cross-modal inconsistencies through intrasentential similarity, so as to lock suspicious tokens. Then, the asymmetric sparsity constraint is proposed to alleviate label noise interference, while the semantic consistency constraint is devised to enhance the reliability of intrasentential visual deviation.
The main contributions are presented as follows:
\begin{itemize}
	\item We explore a new task, image-text multimodal weakly-supervised manipulation localization, and propose the CIEC framework to effectively couple implicit unimodal cues with explicit cross-modal guidance.
	\item We devise the TRPS module for the image WFL subtask. It integrates dual-branch cues via prior-based candidate boxes while suppressing irrelevant responses in non-locked regions.
	\item We propose the VCTG module for the textual WFL subtask. It leverages relative visual bias to assist in verifying token authenticity while mitigating label and cross-modal noise interference. 
\end{itemize}

\section{Methodology}
\subsection{Overview}
\begin{figure*}
	\centering
	{\includegraphics[width=\linewidth]{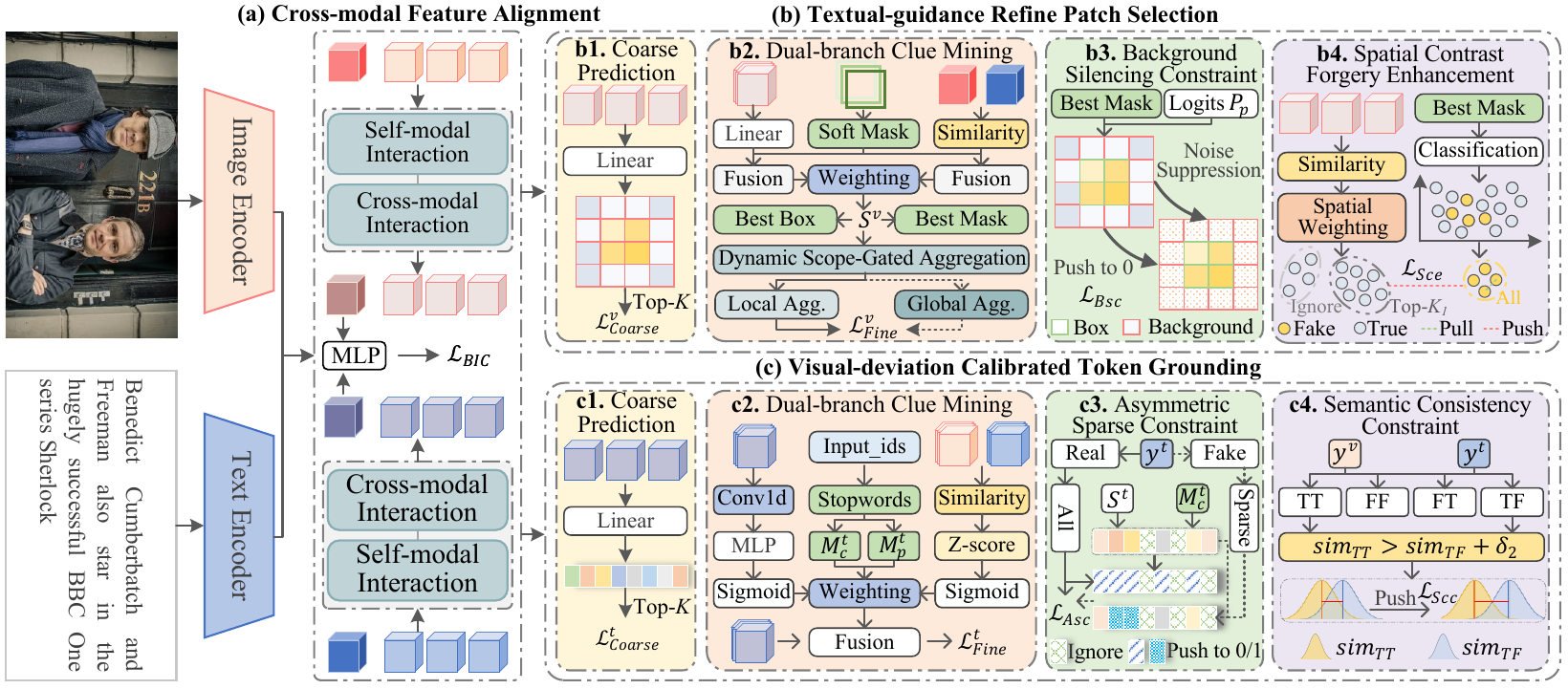}}
	\caption{The overall architecture of proposed CIEC. It consists of three components: (a) Cross-modal Feature Alignment, (b) Textual-guidance Refine Patch Selection and (c) Visual-deviation Calibrated Token Grounding.}
	\label{fig2}
\end{figure*}

To alleviate the heavy burden of expensive fine-grained annotations, we investigate the image-text multimodal weakly-supervised manipulation localization task, aiming to precisely locate forged content relying solely on coarse-grained labels. The overall pipeline is depicted in Figure \ref{fig2}.

Specifically, given a batch of image-text pairs, we first feed them into the CFA module. Here we first extract the unimodal features denoted as $V = [V_{cls}, V_{pat}]$ and $T = [T_{cls}, T_{tok}]$, utilizing their respective encoders. Where $V_{cls}$ and $T_{cls}$ denote the global features, $V_{pat}$ and $T_{tok}$ denote the local features. Subsequently, we feed $V$ and $T$ into the feature interaction layer to obtain enhanced representations, denoted as $\hat{V} = [\hat{V}_{cls}, \hat{V}_{pat}]$ and $\hat{T} = [\hat{T}_{cls}, \hat{T}_{tok}]$.

Second, these features are fed into the TRPS module for image WFL. Here, we first utilize Top-$K$ algorithm to aggregate $\hat{V}_{pat}$ and obtain the coarse-grained prediction. Subsequently, we refine the candidate boxes with the highest forgery probability via the implicit-explicit dual-branch cue mining layer, while simultaneously employing a dynamic score-gated aggregation mechanism to achieve fine-grained prediction. Furthermore, we propose a background silencing constraint to mitigate interference from non-locked regions. And then we employ a spatial contrast forgery enhancement layers to further amplify the distinction between true and fake regions.

Third, the aforementioned features are fed into the VCTG module for textual WFL. Here, we similarly utilize Top-$K$ algorithm to aggregate $\hat{T}_{tok}$ to obtain the coarse-grained prediction. Subsequently, we min the intrinsic textual forgery cues and extrinsic cross-modal inconsistency cues  to achieve fine-grained prediction. Furthermore, we propose an asymmetric sparse constraint to constrain the objective fact that ``each token in a authentic sentence is true'' and ``A fake sentence must contain several fake tokens.'' And then we employ a semantic consistency constraint to enhance the reliability of cross-modal clue extraction.

Finally, we combine all losses via constant-weighted aggregation and feed it into the optimizer for training.

\subsection{Cross-modal Feature Alignment}

To bridge the semantic gap and heterogeneity between the two modalities, we design the CFA module consisting of $N_1$ stacked co-attention layers. This module adopts a dual-stream architecture, allowing the visual and textual features to guide each other's refinement process symmetrically.

Specifically, we first utilize the image encoder to extract raw visual features denoted as $V = [V_{cls}, V_{pat}]$, where $V_{cls} \in \mathbb{R}^{D}$, $V_{pat} \in \mathbb{R}^{N \times D}$, $N$ denotes the numbers of patch and $D$ denotes the dimensions. Similarly, we utilize the text encoder to extract raw textual features denoted as $T = [T_{cls}, T_{tok}]$, where $T_{cls} \in \mathbb{R}^{D}$, $T_{tok} \in \mathbb{R}^{L \times D}$ and $L$ denotes the numbers of token.

Then, these features are fed into two identical but separate processing pipelines comprising the self-modal and the cross-modal interaction layer.
After $N_1$ layers of interaction, we can obtain $\hat{V} = [\hat{V}_{cls}, \hat{V}_{pat}] $ and $\hat{T} = [\hat{T}_{cls}, \hat{T}_{tok}] $, which are implicitly aligned in the semantic space and ready for the subsequent fine-grained localization tasks.

As a result, the multimodal binary classification loss is calculated by
\begin{equation}
	\mathcal{L}_{BIC} = \mathrm{BCE} \left( \mathrm{MLP} \left(\mathrm{Concat} \left( \hat{V}_{cls}, \hat{T}_{cls}   \right)  \right)
	, y^m  \right)
\end{equation}
where $\mathrm{BCE}$ denotes the Binary Cross-Entropy. $\mathrm{MLP}$ consists of linear layers with a hidden expansion ratio of 2, interleaved with layer normalization and GELU activation function. $y^m\in \{0,1\}$ denotes the multimodal ground-truth.

\subsection{Textual-guidance Refine Patch Selection}
Although existing image WFL \cite{zhai2023towards, zhou2024exploring, sheng2025weakly, li2025m2rl} have achieved notable progress, they suffer from severe performance degradation on the DGM$^4$ dataset \cite{shao2023detecting} due to the extreme sparsity of manipulation traces. Specifically, manipulated regions in previous datasets \cite{dong2013casia, wen2016coverage} typically exceed 10\% of the image area, whereas in DGM$^4$, nearly half of the instances fall below 5\%. Such subtle manipulations cause prior global attention-based methods to be easily overwhelmed by background noise, rendering WFL ineffective.

Considering that manipulations in DGM$^4$ primarily target the facial region, where the adverse effects are most severe, we propose a candidate box-assisted localization paradigm. Here, instead of conducting a blind search across the entire image, we leverage existing face detectors to extract candidate boxes as spatial priors. This significantly reduces the complexity of the search space and effectively filters out interference from irrelevant backgrounds.

Based on this premise, we construct the TRPS module, which aims to integrate dual-branch cues consisting of implicit visual anomalies and explicit cross-modal conflict to select the best proposal, and then utilizes two constraints to enhance discriminability.

\subsubsection{Coarse Prediction}
This layer is designed to supervise the accuracy of local cues utilizing only image-level labels. Specifically, we first project the visual features $\hat{V}_{\text{pat}}$ into a patch-level anomaly score map. Subsequently, considering the sparsity of forged regions, we employ a Top-$K$ mean pooling strategy to calculate the image-level forgery prediction score denoted as $\hat{y}^v_{coarse}$.
Finally, the BCE is utilized to encourage the model to focus on the most discriminative visual regions, thereby establishing a solid feature foundation for the subsequent text-guided refinement,
\begin{equation}
	\mathcal{L}^v_{Coarse} = \mathrm{BCE} \left( \hat{y}^v_{coarse}, y^v  \right)
\end{equation}
where $y^v \in \{0,1\}$ denotes the image-level ground-truth.

\subsubsection{Dual-Branch Clue Mining}
This layer aims to mine forgery traces from the dual perspectives of implicit visual anomalies and explicit cross-modal conflicts, thereby identifying the most suspicious proposal based on the prior candidate boxes.

Specifically, given a batch of image-text pairs, there has a batch corresponding candidate proposals denoted as $\mathcal{C} = \{\mathcal{C}_1, \dots, \mathcal{C}_B\} \in \mathbb{R}^{B \times n \times 4}$, where $B$ denotes the batch size, $\mathcal{C}_i = \{C_1, \dots, C_n\} \in \mathbb{R}^{n \times 4}$ denotes the $n$ candidate boxes corresponding to the $i$-th image. Here we first generate the corresponding soft mask. Notably, unlike traditional hard cropping, e.g., ROI Pooling, which is non-differentiable with respect to box coordinates, we adopt a Sigmoid-based soft masking strategy to ensure gradient propagation. Intuitively, we compute the signed distance from each patch center to the candidate box boundaries, where patches inside the box yield positive distances and mask values close to 1, while those outside yield negative distances and decay smoothly to 0. 

Take $\mathcal{C}_i = \{C_1, \dots, C_n\}$ as an example, where $C_i = (c_x, c_y, w, h)$ denotes the normalized center coordinates, width, and height of the $i$-th box, $\left( p_x^{(j)}, p_y^{(j)} \right)$ denotes the grid coordinates of the $j$-th patch scaled to the image size $N$. The soft mask is formulated as the intersection of two smooth box functions along the horizontal and vertical axes. Specifically, the soft mask $M_{i,j}$ for the $j$-th patch w.r.t. box $C_i$ is defined as
﻿\begin{equation}
	\begin{split}
		M_{i,j} = &\sigma\left(\tau_1 \left({N} \cdot w/2 - \left|p_x^{(j)} - N \cdot c_x\right|\right)\right) \\
		\cdot &\sigma\left(\tau_1 \left({N} \cdot h/2 - \left|p_y^{(j)} - N \cdot c_y\right|\right) \right)
	\end{split}
\end{equation}
where $\sigma(\cdot)$ is the Sigmoid function, and $\tau_1 = 2.0$ is a temperature hyperparameter controlling the sharpness of the box boundaries. By repeating this process $n$ times, we can obtain $\mathcal{{M}}_i = \{ {M}_1, \dots, M_n\} \in \mathbb{R}^{n \times N}$. As a result, $\mathcal{M} = \{ \mathcal{M}_1, \dots, \mathcal{M}_B \} \in \mathbb{R}^{B \times n \times N} $ corresponding to $\mathcal{C}$ is derived.

Considering that all prior boxes in the authentic images are incorrect, we filter out this interference during the training phase based on $y^v$, i.e., $ \mathcal{\hat{M}}=\mathcal{M} \cdot y^v$.

Second, we dig up forged clues from each candidate box through two parallel branches. As for implicit visual branch, we first project $\hat{V}_{pat}$ to obtain the patch prediction scores $P_p \in \mathbb{R}^{B \times N}$. And then we calculate the visual suspiciousness by $S^{iv} = \mathrm{Sum} \left ( {\mathcal{\hat{M}} \cdot P_p}  \right )$.
As for explicit semantic branch, we first project original features $V_{pat}$ and $T_{cls}$ into a shared semantic space. And then we calculate the semantic suspiciousness via $S^{ev} =  \mathrm{Sum} \left(  {\mathcal{\hat{M}} \cdot \mathrm{Sim} \left( V_{pat}, T_{cls} \right)} \right)$.
Thus, we can get the final selection score by the weighted fusion of both cues,
\begin{equation}
	S^v = S^{iv} + \alpha \cdot S^{ev}
\end{equation}
where $\alpha$ is a learnable parameter balancing the two modalities. Then, we select the candidate with the maximum $S^v$ as the best proposal, denoted as $\mathcal{C}^* \in \mathbb{R}^{B \times 4}$, where the corresponding mask is denoted as $\mathcal{M}^* \in \mathbb{R}^{B \times N}$.

Third, we propose a dynamic scope-gated aggregation mechanism to translate the local forgery cues into a global image-level prediction, which dynamically switches the aggregation scope (local or global) based on $y^{v}$ in training (or $\hat{y}^v_{coarse}$ in testing). Formally, let $\mathbb{I}_{valid}$ be an indicator function that evaluates to 1 if $y^{v}=1$ (or $\hat{y}^v_{coarse}>0.5$), and 0 otherwise, we have 
\begin{equation}
	\hat{y}^v_{fine} = \mathbb{I}_{valid} \cdot \mathcal{A}_{local} + (1 - \mathbb{I}_{valid}) \cdot \mathcal{A}_{global}
\end{equation}
where $\mathcal{A}_{local}$ and $\mathcal{A}_{global}$ represent the aggregated scores from the local and global scopes, respectively. Specifically, in local aggregation, we consider the scenario where the image is deemed forged. Here we focus on features within the candidate box, utilizing the Log-Sum-Exp (LSE) pooling to smoothly approximate the max operator while maintaining gradient flow, 
\begin{equation}
	\mathcal{A}_{local} = \tau_2 \log \left( \frac{1}{n} \sum_{i=1}^{n} \exp( \left( S^v \right)^i / \tau_2) \right)
\end{equation}
where $\tau_2=0.1$. Note that since $S^v$ is the result of soft mask weighting, it is more sensitive to features within the box. In global aggregation, we consider the opposite scenario. Here, we treat each patch equally, calculating $\mathcal{A}_{global}$ using $P_p$,
\begin{equation}
	\mathcal{A}_{global} = \tau_2 \log \left( \frac{1}{N} \sum_{i=1}^{N} \exp( \left( P_p \right)^i / \tau_2) \right)
\end{equation}

Finally, the fine-grained prediction branch is optimized by
\begin{equation}
	\mathcal{L}^v_{Fine} = \mathrm{BCE} \left( \hat{y}^v_{fine}, y^v  \right)
\end{equation}

\subsubsection{Background Silencing Constraint}
Under WFL settings, the model is prone to attention diffusion, i.e., often falsely activating irrelevant salient regions (e.g., complex but authentic backgrounds surrounding faces). To mitigate this, we introduce the background silencing constraint based on the spatial exclusivity assumption, where all regions outside the best candidate box are considered true. Notably, this constraint is exclusively applied to fake images during training. 

Specifically, based on $\mathcal{M}^*$ corresponding to the best candidate box $\mathcal{C}^*$ selected in the previous stage, we define a binary background indicator $I_{bg}$ to identify background regions,
\begin{equation}
	\label{eq14}
	I_{bg} = \mathbb{I} \left( \mathcal{M}^*  < \epsilon \right)
\end{equation}
where $\epsilon = 0.1$ denotes the segmentation threshold.
Then, we employ a weighted BCE loss with a zero target to suppress background,
\begin{equation}
	\mathcal{L}_{Bsc} = \mathrm{BCE} \left( P_p, 0, \text{weight} = I_{bg}  \right)
\end{equation}
By minimizing this loss, the model can more clearly distinguish between true background and fake regions.

\subsubsection{Spatial Contrast Forgery Enhancement}
Although $\mathcal{L}_{Bsc}$ mitigates noise interference outside candidate regions, it fails to truly distinguish between true and fake regions. This leads to confusion when the background outside the box closely resembles the forged cue, prompting the model to question, “Why ignore something that resembles the forged cue?” Consequently, it reduces loss by forcibly distorting classifier weights instead of optimizing the feature extractor to minimize loss. To fundamentally enhance discriminative capability, we propose the spatial contrast forgery enhancement layer, which aims to pull closer fake patches while pushing away fake patches from highly similar true patches, thereby enabling the model to clearly distinguish between true and fake regions.

Specifically, we first partition the patch set into a forged subset $\mathcal{P}_{Fake}$ and a true background subset $\mathcal{P}_{True}$, based on \eqref{eq14}. Then, we compute the pairwise cosine similarity matrix over all patches utilizing $\hat{V}_{pat}$. Second, we further modulate similarity matrix utilizing the Euclidean distance to encourage local consistency through spatial distance, where closer distances carry greater similarity weight, thereby obtaining the spatially weighted similarity matrix $\mathit{H}$. Third, we introduced two constraints: 1) For both patches are fake, we encourage their representations to be compact and consistent. 2) For one true and one fake, we adopt hard negative mining to increase separability, where we encourage the Top-$K_1$ most similar pairs to exhibit greater inconsistency in representation. Finally, the supervision is implemented with a margin ranking loss,
\begin{equation}
	\mathcal{L}_{Sce} = \mathrm{ReLu} \left( \delta_1 - \mathit{H}^{(ij)} \right) + \mathrm{ReLu} \left(\delta_1 + \mathit{H}^{(ik)} \right)
\end{equation}
where $i,j \in \mathcal{P}_{Fake}$, $k \in \text{Top-}K_1(\mathcal{P}_{True})$, $\delta_1 =0.2$ denotes the margin. Based on this, we can effectively sharpen the feature boundary, ensuring that the mined forgery traces are distinct from complex background textures.

\subsection{Visual-deviation Calibrated Token Grounding}
This module aims to achieve textual WFL under visual feature calibration. Specifically, it integrates a dual-branch clue consisting of intrinsic textual forgery and extrinsic cross-modal inconsistency to select fake tokens, and then utilizes asymmetric sparse and semantic consistency constraints to enhance reliability.

\subsubsection{Coarse Prediction}
The execution process of this layer is consistent with its counterpart in the visual branch. Here, we also employ the Top-$K$ mean pooling strategy to obtain coarse-grained predictions denoted as $\hat{y}^t_{coarse}$, followed by the BCE to optimize feature representations,
\begin{equation}
	\mathcal{L}^t_{Coarse} = \mathrm{BCE} \left( \hat{y}^t_{coarse}, y^t  \right)
\end{equation}
where $y^t \in \{0,1\}$ denotes the text-level ground-truth label.

\subsubsection{Dual-Branch Clue Mining}
In fact, forgery traces in text are extremely subtle and exceptionally difficult to detect. Especially in WFL, achieving precise localization solely through textual features is even more challenging. To address this, we propose mining forgery traces from two perspectives: intrinsic textual forgery and extrinsic cross-modal inconsistency.

Specifically, we first generate two masks based on the original annotations, a padding mask denoted as $M^t_p$ and a stop-word mask denoted as $M^t_c$ that masks meaningless words. Second, for the intrinsic textual forgery mining branch, we utilize a 1D convolutional network with a kernel size of 3 to capture intrinsic manipulation clue such as incoherent phrasing,
\begin{equation}
	S^{it} = M^t_c \cdot \sigma \left(
	\text{Linear} \left( \text{Conv1d} \left( \hat{T}_{tok}
	\right) \right) \right)
\end{equation}
Note that here we focus solely on meaningful content words, as they pose a greater risk. For the extrinsic cross-modal inconsistency mining branch, we first compute the similarity between each token and all patches utilizing original features, and then find the best-matched visual patch for each token as its visual support clue, 
\begin{equation}
	\label{eq18}
	S_{raw} = \max_{i} \left( \mathrm{Sim} \left(
	T_{tok}, V_{pat}^{(i)}
	\right) \right)
\end{equation}
Given the significant variability in similarity scores across different image-text pairs, we employ the in-sentence Z-score normalization to calibrate $S_{raw}$ and then map it to capture extrinsic inconsistency clue,
\begin{equation}
	S^{et} = M^t_c \cdot \sigma \left( \eta  \beta - \eta  \Upsilon \left(
	{S}_{raw}
	\right) \right)
\end{equation}
where $\eta$ and $\beta$ are learnable scale and center parameters. $\Upsilon(\cdot)$ denotes in-sentence Z-score normalization, which is achieved by simple variance estimation. Note that a lower similarity leads to a higher forgery probability.

Second, the token-level forgery predictions from both branches are adaptively fused via learnable gating weights,
\begin{equation}
	S^{t} = W^{it} \cdot S^{it} + W^{et} \cdot S^{et}
\end{equation}
Finally, we derive sentence-level prediction scores $\hat{y}^t_{fine}$ by performing a weight-sum calculation for $\hat{T}_{tok}$ utilizing $S^{t}$. Then, it is optimized by
\begin{equation}
	\mathcal{L}^t_{Fine} = \mathrm{BCE} \left( \hat{y}^t_{fine}, y^t  \right)
\end{equation}

\subsubsection{Asymmetric Sparse Constraint}

To achieve token-level localization under limited sentence-level weakly-supervised labels, we propose the asymmetric sparse constraint. This constraint is grounded in the linguistic premise that every token in an authentic sentence is true, whereas a forged sentence typically contains only sparse manipulated keywords (e.g., entities or verbs). Specifically, we formulate the training process into two asymmetric branches: hard suppression for true samples and sparse activation for fake samples.

For the former ($y^t=0$), we enforce the forgery probabilities of all valid tokens to approach zero, thereby minimizing false positives on background words,
\begin{equation}
	\mathcal{L}_{Asc}^{T} = \underset{{i \in \Omega_{True}}}{\mathbb{E}} \left[  \mathrm{BCE} \left( \left( S^{t} \right)^{(i)}, 0 \right) \right] 
\end{equation}
where $\Omega_{True}$ denotes the set of true sentences.

For the latter ($y^t=1$), distinct from the standard MIL approaches that assume all instances in the fake samples contribute to the label, we only assume partial instances contribute to the label. This assumption mitigates interference from label noise (e.g., erroneously treating unimportant function words as fake), so as to enhance detection capabilities under sparse manipulation localization. Here, we dynamically determine the number of sparse activations based on sentence length and select the Top-$K_2$ tokens with the highest forgery probability as the index set $O$. Subsequently, we optimize the loss only on these selected tokens,
\begin{equation}
	\mathcal{L}_{Asc}^{F} = \underset{{i \in \Omega_{Fake}, j \in O}}{\mathbb{E}} \left[  \mathrm{BCE} \left( \left( S^{t} \right)^{(ij)}, 1 \right) \right] 
\end{equation}
Additionally, for tokens not belonging to $O$, we impose no operations on them. This prevents the model from overfitting irrelevant function words in fake sentences, thereby robustly handling the sparsity of manipulation traces.

Finally, the full asymmetric sparse constraint loss is derived by
\begin{equation}
	\mathcal{L}_{Asc}=\lambda \cdot \mathcal{L}_{Asc}^{T}+\mathcal{L}_{Asc}^{F}
\end{equation}
where $\lambda = 2.0$, is used to forcefully suppress noise in authentic samples to ensure precision.

\begin{table*}[t]
	\small
	\centering
	\caption{Comparison with SOTA methods on the entire DGM$^4$. $\downarrow$ means less is better.}
	\begin{tabular}{ccllccccccccc}
		
		\hline\hline
		& & \multirow{2}{*}[-0.5ex]{Method} & \multirow{2}{*}[-0.5ex]{Reference} & \multicolumn{3}{c}{Binary Cls} & \multicolumn{3}{c}{Image Grounding} & \multicolumn{3}{c}{Text Grounding} \\ \cmidrule(lr){5-7} \cmidrule(lr){8-10} \cmidrule(lr){11-13}
		& & &                       & AUC      & EER$\downarrow$      & ACC   &IoU$_m$     & IoU50     & IoU75     & Precision    & Recall    & F1 \\ \hline
		
		\multirow{12}{*}{\rotatebox{90}{\textbf{Entire Dataset}}}
		& \multicolumn{1}{c}{\multirow{9}{*}{{\rotatebox{90}{Fully}}}} 
		
		& CLIP  & ICML21 & 83.22 & 24.61 & 76.40 & 49.51 & 50.03 & 38.79 & 58.12 & 22.11 & 32.03 \\
		& & ViLT & ICML21 & 85.16  & 22.88 & 78.38 & 59.32 & 65.18 & 48.10 & 66.48 & 49.88 & 57.00 \\ 
		& & HAMMER  & CVPR23 & 93.19 & 14.10 & 86.39 & 76.45 & 83.75  & 76.06 & 75.01        & 68.02     & 71.35 \\ 
		& & HAMMER++  & TPAMI24 &93.33    & 14.06    & 86.66   & 76.46       & 83.77     & 76.03     & 73.05        & 72.14     & 72.59 \\ 
		& & ViKI   & IF24 & 93.51    & 13.87    & 86.67  & 76.51       & 83.95     & 75.77     & 77.79        & 66.06     & 72.44 \\ 
		& & UFAFormer & IJCV24 & 93.81    & 13.60    & 86.80  & 78.33       & 85.39     & 79.20     & 73.35        & 70.73     & 72.02 \\ 
		& & IDseq & AAAI25 & 94.55    & 11.40    & 88.94  & 83.33       & 89.39     & 86.10     & 75.96        & 71.23     & 73.52 \\ 
		& & ASAP & CVPR25 & 94.38    & 12.73    & 87.71   & 77.35       & 84.75     & 76.54     & 79.38        & 73.86     & 76.52 \\
		& & CSCL  & CVPR25  & 96.34  & 9.88  &90.32  &84.07  &90.48  &87.17    & 75.33   & 77.95    & 76.62	\\ \cline{2-13}
		& \multicolumn{1}{c}{\multirow{3}{*}{{\rotatebox{90}{Weakly}}}}
		& HAMMER-w & CVPR23 &91.52  &16.25  &84.14  &63.18  &66.33  &65.23  &38.94  &64.03  &48.43  \\
		& & CSCL-w & CVPR25  &95.41  &10.95  &89.33  &67.94  &71.41  &70.23  &46.78  &58.17  &51.86 \\ 
		& & \cellcolor{gray!20}CIEC (Ours) & \cellcolor{gray!20}-  & \cellcolor{gray!20}\textbf{95.67} & \cellcolor{gray!20}\textbf{10.76}  & \cellcolor{gray!20}\textbf{89.40} & \cellcolor{gray!20}\textbf{83.49} & \cellcolor{gray!20}\textbf{88.94} & \cellcolor{gray!20}\textbf{87.12} & \cellcolor{gray!20}\textbf{53.24} & \cellcolor{gray!20}\textbf{60.85} & \cellcolor{gray!20}\textbf{56.80} \\ \hline
		\hline
	\end{tabular}
	\label{tab1}
\end{table*}

\subsubsection{Semantic Consistency Constraint}
As stated above, we utilize relative visual deviation to assist token localization. Yet it relies on a premise that authentic image-text pairs exhibit stronger similarity compared to manipulated ones. To explicitly enforce this distributional disparity and enhance the reliability of the visual branch, we propose the semantic consistency constraint.

Specifically, we first filter the current batch based on $y^t$ to form two sets: pure true $\mho_1$ and true image with fake text $\mho_2$. Notably, here we exclude visual forgeries to prevent feature contamination. Next, we compute the average similarity of all valid tokens (excluding padding) within each set utilizing Eq. \eqref{eq18}. Finally, a margin ranking loss is imposed to ensure that the semantic affinity of genuine pairs is statistically significantly higher than that of manipulated pairs,
\begin{equation}
	\mathcal{L}_{Scc} = \mathrm{ReLu} \left( \delta_2 - M^p_c \cdot \mathbb{E} \left[ (S_{raw})^{(i)} - (S_{raw})^{(j)} \right] \right) 
\end{equation}
where $\delta_2 = 0.1$, $i \in \mho_1$ and $j \in \mho_2$. By optimizing $\mathcal{L}_{Scc}$, we ensure that manipulated texts are statistically pushed to the lower tail of the similarity distribution, thereby boosting the reliability of the visual-aid branch.

\section{Experiments}
\subsection{Dataset and Metrics}
Consistent with all SOTA methods \cite{wang2024exploiting, liu2025idseq, zhang2025asap}, our experiments are carried out on the DGM$^4$ dataset \cite{shao2023detecting}, which employs four forgery methods: Face Swap (FS), Face Attribute (FA), Text Swap (TS), and Text Attribute (TA), covering eight types of multimodal manipulation.

For performance evaluation, we follow the same evaluation metrics adopted in prior studies \cite{liu2025idseq, zhang2025asap}. Notably, precisely identifying forgery methods (such as FS and FA) is extremely challenging when only coarse-grained classification labels are readily available. Therefore, this paper omits this difficult task.

\subsection{Implementation Details}
The input image is resized to $256 \times 256$, and the text sequence is padded to a length of 50 tokens. Then, we utilize ViT-B/16 \cite{dosovitskiy2020image} and RoBERTa \cite{liu2019roberta} to encode image and text, with pre-trained backbone weights sourced from METER \cite{dou2022empirical}. More specifically, the cross-modal interaction layer $N_1$ is set to 6. The prior facial candidate boxes are obtained by MTCNN \cite{zhang2016joint}, and the number of candidate boxes $n$ is set to 5. 
For optimization, we utilize AdamW with a weight decay of 0.02, while the learning rate is set to $1\times 10^{-5}$. The proposed model is trained for 50 epochs on 8 A100 GPUs with a batch size of 32.

\subsection{Compared with State-of-the-Art Methods}
We compare our CIEC with nine SOTA fully supervised methods \cite{radford2021learning, kim2021vilt, shao2023detecting, shao2024detecting, li2024towards, liu2024unified, liu2025idseq, zhang2025asap, li2025unleashing}. Furthermore, given that currently no multimodal WFL methods exist for image-text pairs manipulation, we modify two fully supervised methods \cite{shao2023detecting, li2025unleashing} into weakly-supervised variants to serve as fair baselines.

Specifically, we first remove the layers dependent on patch-level or token-level annotations from the original network. Then, we add the candidate-box-based visual-branch clue mining layer ($S^{iv}$) from CIEC to achieve image WFL. Third, we add the intrinsic semantic mining layer ($S^{it}$) and asymmetric sparse constraint ($\mathcal{L}_{Asc}$) from CIEC for textual WFL. For all other designs, we retain the original details, ensuring a fair comparison under the same setting.

As shown in Table \ref{tab1}, despite the lack of fine-grained supervision, CIEC exhibits remarkable performance, comparable to or even surpassing fully supervised methods. Specifically, for the binary detection subtask, CIEC improves the AUC by 2.48\%,1.86\% and 1.29\%, on HAMMER \cite{shao2023detecting}, UFAFormer \cite{liu2024unified} and ASAP \cite{zhang2025asap}, respectively. It is slightly lower than CSCL \cite{li2025unleashing}, as CSCL has more fine-grained annotations, including multi-class and fine-grained localization annotations, which aid in learning the binary detection subtask.
For the image grounding subtask, CIEC gains +11.35\%, +1.02\% and +10.58\% IOU75 on ViKI \cite{li2024towards}, IDseq \cite{liu2025idseq} and ASAP, respectively. This demonstrates that the proposed TRPS effectively mines discriminative visual forgery cues even under coarse supervision.
For the text grounding subtask, CIEC achieves 56.80\% F1 score. Although it performs on par with ViLT \cite{dosovitskiy2020image}, a gap remains compared to SOTA methods. This is because TS and TA manipulation leave few forgery traces in the text \cite{liu2024unified, yu2025fine}. Particularly under WFL settings, localizing forgery clues within the text becomes even more challenging.

Furthermore, compared to the weakly-supervised methods, CIEC demonstrates absolute superiority, particularly in image localization, achieving at least a 12\% improvement over existing SOTA methods. Specifically, CIEC outperforms HAMMER-w \cite{shao2023detecting} by 4.15\% AUC, 5.26\% ACC, 22.61\% IoU50 and 8.37\% in F1. Besides, CIEC gains +16.06\% IoU75 and 4.96\% F1 on SOTA CSCL-w \cite{li2025unleashing}.
These consistent improvements indicate that our coupled implicit-explicit cue mining strategy extracts forgery evidence much more accurately than naive adaptations of existing frameworks.

\begin{figure*}[!t]
	\centering
	{\includegraphics[width=\linewidth]{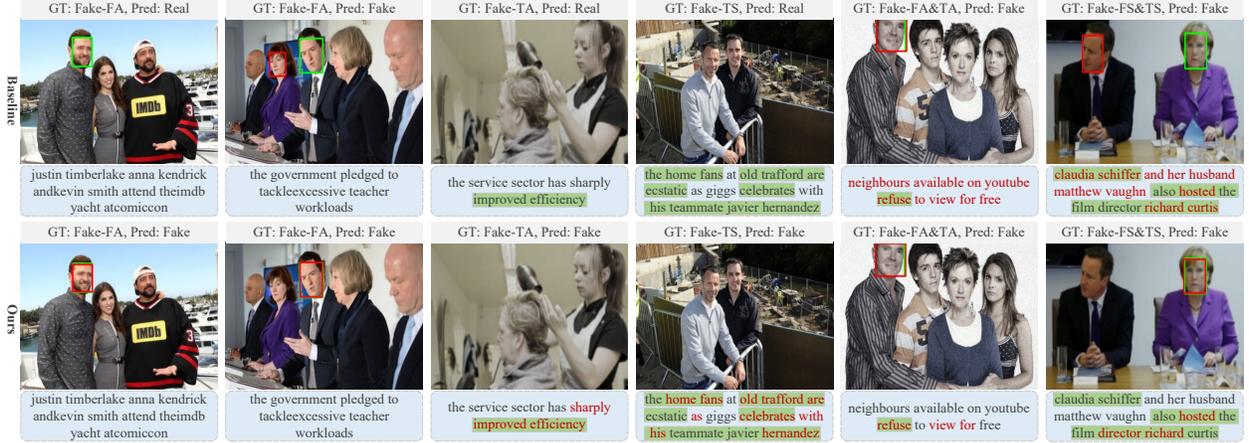}}
	\caption{Visualization of detection and grounding results. Here, red box and text indicate the prediction of manipulated faces and words, while green box and text represent the corresponding ground truth.}
	\label{fig3}
\end{figure*}

\subsection{Ablation Study}

We first conduct an ablation experiment on the TRPS module. As presented in Table \ref{tab3}, the complete method yields the best results. Specifically, when $\mathcal{L}_{Fine}^v$ is omitted, the localization capability plummets dramatically, which confirms that our dual-branch clue mining layer is the cornerstone for handling the sparse manipulation traces. 
When $\mathcal{L}_{Bsc}$ is excluded, IoU75 drops 8.08\%, indicating that the model struggles to suppress irrelevant background noise without this explicit spatial regularization. 
When the textual guidance branch is disabled, IoU$_m$ degrades 0.65\%, which suggests that explicit semantic guidance serves as a vital means to complement implicit visual cues.

Then, we perform a fine-grained ablation on the VCTG module. As shown in Table \ref{tab4}, the complete method is also the best. Specifically, removing $\mathcal{L}_{Asc}$ makes F1 plunge from 56.80\% to 30.89\%.
This confirms that our asymmetric sparse activation strategy effectively achieves forgery localization for meaningful tokens, reducing label noise caused by the full activation of standard MIL.
Moreover, disabling the visual branch decreases 11.34\% F1 score. We attribute this to the fact that linguistic anomalies alone are insufficient to detect semantically plausible yet cross-modal conflicting forgeries, such as TA. 
Similar performance drops can also be observed in other variants.

\begin{table}[t]
	\small
	\setlength\tabcolsep{4.2pt}
	\centering
	\caption{Ablation study on the TRPS module.}
	\begin{tabular}{lccccccc}
		\hline\hline
		\multirow{2}{*}[-0.5ex]{Variant} & \multicolumn{3}{c}{Binary Cls} & \multicolumn{3}{c}{Image Grounding} \\ \cmidrule(lr){2-4} \cmidrule(lr){5-7} 
		& AUC      & EER$\downarrow$      & ACC   &IoU$_m$     & IoU50     & IoU75 \\ \hline
		w/o $\mathcal{L}^v_{Coarse}$  & 95.17 & 89.07 & 11.15 & 82.49 & 87.90 & 86.09 \\
		w/o $\mathcal{L}^v_{Fine}$ & 94.98 & 88.09 & 12.24 & 47.22 & 47.22 & 47.22 \\
		w/o $\mathcal{L}_{Bsc}$ & 94.54 & 87.36 & 13.24 & 76.14 & 80.53 & 79.04 \\
		w/o $\mathcal{L}_{Sce}$ & 95.15 & 89.02 & 11.18 & 82.76 & 88.21 & 86.38 \\
		w/o Text-aid & 95.48 & \textbf{89.40} & 10.79 & 82.84 & 88.38 & 86.51 \\
		Full & \textbf{95.67} & \textbf{89.40} & \textbf{10.76} &\textbf{83.49} & \textbf{88.94} & \textbf{87.12} \\
		\hline\hline
	\end{tabular}
	\label{tab3}
\end{table}

\begin{table}[t]
	\small
	\setlength\tabcolsep{4.5pt}
	\centering
	\caption{Ablation study on the VCTG module. PR. represents Precision, while RE. represents Recall.}
	\begin{tabular}{lccccccc}
		\hline\hline
		\multirow{2}{*}[-0.5ex]{Variant} & \multicolumn{3}{c}{Binary Cls} & \multicolumn{3}{c}{Text Grounding} \\ \cmidrule(lr){2-4} \cmidrule(lr){5-7} 
		& AUC      & EER$\downarrow$      & ACC   & PR.    & RE.    & F1 \\ \hline
		w/o $\mathcal{L}^t_{Coarse}$  &95.47 & 89.29 & 11.13 & 53.59 & 56.44 & 54.98 \\
		w/o $\mathcal{L}^t_{Fine}$ &95.61 & 89.24 & 10.98 & 55.63 & 51.48 & 53.47 \\
		w/o $\mathcal{L}_{Asc}$ &95.61 & \textbf{89.52} & 10.77 & 43.41 & 23.97 & 30.89 \\
		w/o $\mathcal{L}_{Scc}$ &95.31 & 89.43 & 10.79 & 53.41 & 59.67 & 56.37 \\
		w/o Image-aid &\textbf{95.67} & 89.30 & 10.91 & \textbf{59.17} & 36.91 & 45.46 \\
		Full & \textbf{95.67} & 89.40 & \textbf{10.76} & 53.24 & \textbf{60.85} & \textbf{56.80} \\
		\hline\hline
	\end{tabular}
	\label{tab4}
\end{table}

\subsection{Qualitative Results}
Here we implement a baseline by simplifying our CIEC. Specifically, we first directly utilize the features generated by the feature extractor to perform binary classification tasks. Then, we utilize a parallel forgery clue mining layer identical to HAMMER-w and CSCL-w, to achieve image WFL and text WFL, respectively.

Figure \ref{fig3} presents the qualitative comparison results between CIEC and the baseline model across various manipulation types. As illustrated in the top row, the baseline often fails to detect subtle manipulations, frequently leading to false negative predictions or failing to identify the specific manipulated regions. In contrast, as shown in the bottom row, CIEC precisely locates visual forgeries and text manipulations. These results strongly validate the effectiveness of our proposed dual-branch clue mining layer and multiple constraints in capturing fine-grained forgery.

\section{Conclusion}
We propose a novel framework named CIEC to enable precise multimodal manipulation localization, without relying on expensive fine-grained annotations.
It aims to couple implicit monomodal cues with explicit cross-modal guidance to address the challenges of highly independent multimodal weak-supervised tamper localization.
Specifically, TRPS is proposed to achieve image WFL, which utilizes spatial prior to alleviate search complexity and devises background silencing and spatial contrast constraints to suppress interference from irrelevant areas.
Moreover, VCTG is presented to achieve textual WFL, which focuses on meaningful content words and utilizes asymmetric sparse and semantic consistency constraints to ensure cue reliability.
Quantitative and qualitative evaluations on the DGM$^4$ benchmark demonstrate the superiority of our CIEC. Particularly in image grounding subtask, CIEC achieves at least +12\% improvement over all WFL methods. In the future, we will explore leveraging LLMs to assist in mining textual forgery traces.

\bibliography{index_paper}
\bibliographystyle{unsrtnat}

\end{document}